%% file: main.tex
\documentclass[10pt,twocolumn,letterpaper]{article}

\usepackage[pagenumbers]{cvpr} 

\input{preamble}
\definecolor{cvprblue}{rgb}{0.21,0.49,0.74}
\usepackage[pagebackref,breaklinks,colorlinks,allcolors=cvprblue]{hyperref}
\usepackage{multirow}

\def\paperID{} 
\def\confName{CVPR}
\def\confYear{2026}

\title{Mamba Learns in Context: Structure-Aware Domain Generalization\\for Multi-Task Point Cloud Understanding}

\author{
Jincen Jiang$^{1}$ \quad
Qianyu Zhou$^{2\,\dagger}$ \quad
Yuhang Li$^{3}$ \quad
Kui Su$^{4}$ \quad
Meili Wang$^{5}$ \\
Jian Chang$^{1}$ \quad
Jian Jun Zhang$^{1}$ \quad
Xuequan Lu$^{3\,\dagger}$\\[0.5em]
$^{1}$Bournemouth University \quad
$^{2}$Jilin University \quad
$^{3}$The University of Western Australia\\
$^{4}$Hangzhou City University \quad
$^{5}$Northwest A\&F University\\[0.5em]
{\tt\small jiangj@bournemouth.ac.uk \quad zhouqianyu@jlu.edu.cn \quad bruce.lu@uwa.edu.au}\\
}

\begin{document}
\maketitle

\begingroup
\renewcommand\thefootnote{}
\footnotetext{$^\dagger$Corresponding authors.}
\endgroup

\input{sections/0_abstract}    
\input{sections/1_introduction}
\input{sections/2_related_work}
\input{sections/3_method}
\input{sections/4_experiments}
\input{sections/5_conclusion}

\section*{Acknowledgments}
This work is supported by the National Natural Science Foundation of China (No. 62502178), Jilin University Special Program for Talent Development in Engineering Cluster Construction, the China Scholarship Council (No. 202306300023), and the Research and Development Fund of Bournemouth University.

\small
\bibliographystyle{ieeenat_fullname}
\bibliography{main}

\input{sections/X_suppl}

\end{document}

%% file: sections/0_abstract.tex
\begin{abstract}
While recent Transformer and Mamba architectures have advanced point cloud representation learning, they are typically developed for single-task or single-domain settings. Directly applying them to multi-task domain generalization (DG) leads to degraded performance. Transformers effectively model global dependencies but suffer from quadratic attention cost and lack explicit structural ordering, whereas Mamba offers linear-time recurrence yet often depends on coordinate-driven serialization, which is sensitive to viewpoint changes and missing regions, causing structural drift and unstable sequential modeling. In this paper, we propose Structure-Aware Domain Generalization (SADG), a Mamba-based In-Context Learning framework that preserves structural hierarchy across domains and tasks. We design structure-aware serialization (SAS) that generates transformation-invariant sequences using centroid-based topology and geodesic curvature continuity. We further devise hierarchical domain-aware modeling (HDM) that stabilizes cross-domain reasoning by consolidating intra-domain structure and fusing inter-domain relations. At test time, we introduce a lightweight spectral graph alignment (SGA) that shifts target features toward source prototypes in the spectral domain without updating model parameters, ensuring structure-preserving test-time feature shifting. In addition, we introduce MP3DObject, a real-scan object dataset for multi-task DG evaluation. Comprehensive experiments demonstrate that the proposed approach improves structural fidelity and consistently outperforms state-of-the-art methods across multiple tasks including reconstruction, denoising, and registration.
Our source code is available at: https://github.com/Jinec98/SADG.
\end{abstract}

%% file: sections/1_introduction.tex
\section{Introduction}
Understanding 3D point clouds is essential for perception~\cite{qi2017pointnet,qi2017pointnet++,graham20183d,qiu2021dense,milioto2019rangenet++,zhang2020polarnet,wu2019squeezesegv2}, reconstruction~\cite{jenke2006bayesian,berger2014state,huang2024surface,lin2018learning,huang2009consolidation,liu2024cloudmix}, and interaction~\cite{zhen3d,geng2023gapartnet,geng2023partmanip,qin2023dexpoint,li2026pointvla} in real-world systems. Most recent advances build on Transformer-based architectures~\cite{guo2021pct,zhao2021pointtransformer,pang2022masked,lai2022stratified,ando2023rangevit,wu2022point,wu2024point} to capture long-range dependencies via self-attention, while recent work explores State-Space Models such as Mamba~\cite{liang2024pointmamba,zhang2025point,han2024mamba3d,bahri2025spectral,zhang2024voxel,lu2025exploring} for linear-time sequence modeling. Although these architectures achieve strong performance on standard benchmarks, they are typically designed for single-task learning and struggle to generalize across unseen domains or handle multiple point cloud understanding tasks such as reconstruction, denoising, and registration. In practice, sensor variation, viewpoint differences, and scene incompleteness significantly challenge their generalization ability.

To address multi-task domain generalization (DG) for point clouds, DG-PIC~\cite{jiang2024dg} is the first work to explore this direction. It conditions target samples on source-domain prompts, enabling a unified In-Context Learning model to perform multiple tasks. However, DG-PIC relies on Transformers, inheriting high computational complexity and lacking explicit token ordering. A natural replacement is Mamba, but this raises the following challenges: existing Mamba-based methods often depend on coordinate-driven serialization (\emph{e.g.,} Axis Scanning and Hilbert Curves) for global alignment, which are sensitive to viewpoint changes and missing surfaces. Such serializations often break hierarchical object structure, causing unstable state propagation in Mamba and degrading generalization to unseen domains.

The core difficulty lies in multi-task learning: reconstruction, denoising, and registration all rely on preserving structural hierarchy, including global topology (part-whole spatial organization) and local geometric continuity (surface smoothness and curvature). Under domain shifts such as noise, occlusion, and pose variation, coordinate-driven serializations can distort sequence-local neighborhoods and disrupt intrinsic topological and geometric structure, making Mamba's recurrence fragile and feature-based domain alignment cannot ensure structural consistency. Therefore, robust multi-task DG requires explicitly encoding structure-aware token organization, enabling stable sequential modeling and structurally grounded alignment across domains.

In this paper, we propose \textbf{Structure-Aware Domain Generalization (SADG)}, the first Mamba-based In-Context Learning framework for multi-task point cloud domain generalization. Our core idea is to explicitly serialize and align intrinsic geometric structure across domains and tasks. Our SADG consists of three key components. Firstly, we introduce a \textbf{Structure-Aware Serialization (SAS)} strategy based on two intrinsic spectra: a \textit{Centroid Distance Spectrum} that preserves global topology and a \textit{Geodesic Curvature Spectrum} that captures surface continuity, producing transformation-invariant and structure-consistent token sequences for Mamba that allow recurrent state propagation to reflect the underlying object hierarchy. Secondly, we design \textbf{Hierarchical Domain-Aware Modeling (HDM)} that first consolidates intra-domain structure and then performs inter-domain relational fusion within a unified sequence. Finally, a lightweight test-time \textbf{Spectral Graph Alignment (SGA)} module conducts graph spectral shifting to match target features to source prototypes without model updates, ensuring structure-preserving generalization.

Existing multi-domain multi-task point cloud benchmarks are limited in scale and real-scene variability, particularly in pose, occlusion, and sensor noise. To address this, we introduce \textbf{MP3DObject}, a new dataset of object-level real scans from Matterport3D~\cite{Matterport3D} indoor scans, offering a strong testbed for synthetic-to-real generalization and a valuable resource for broader 3D understanding tasks. Extensive experiments on multiple datasets including this one show our method achieves state-of-the-art results. 

\noindent{Our main contributions are as follows:}
\begin{itemize}
    \item We identify the structural drift challenge in multi-task point cloud DG and present a  Structure-Aware Domain Generalization framework that jointly preserves global topology and local geometry across domains and tasks.
    \item We propose structure-aware serialization for topology and curvature ordering, hierarchical domain-aware modeling for stable cross-domain reasoning, and spectral alignment for test-time structure-preserving shifting.
    \item We introduce MP3DObject, a new object-level dataset derived from Matterport3D, providing diverse real-world scans and a challenging benchmark for evaluating generalization from synthetic to real domains.
\end{itemize}

%% file: sections/2_related_work.tex
\begin{figure*}[htbp]
    \centering
    \includegraphics[width=0.91\textwidth]{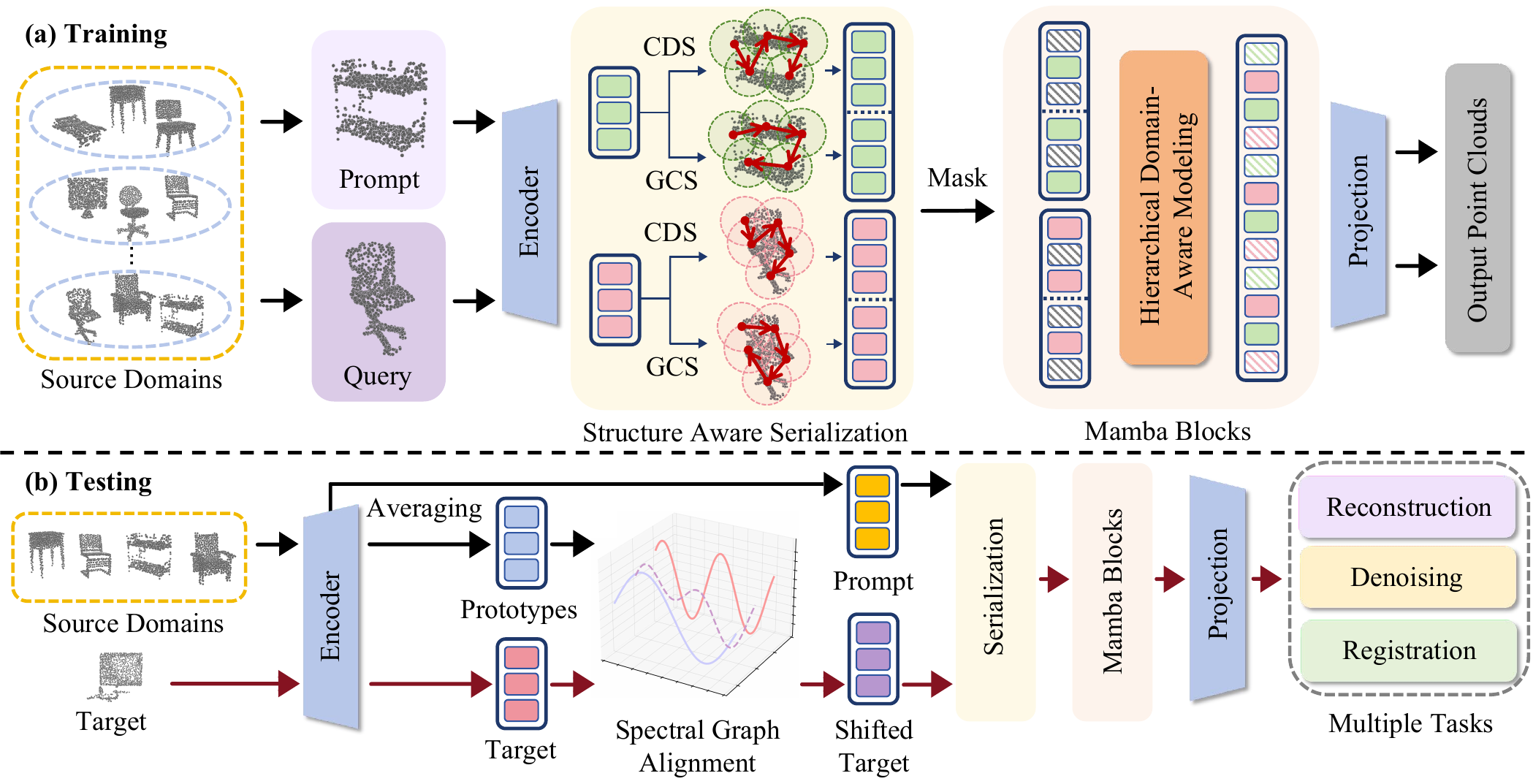}
    \caption{Overview of our \textbf{SADG}. (a) During training, point clouds from multiple source domains are partitioned into local patches and serialized into structure-aware sequences using the Centroid Distance Spectrum (CDS) and Geodesic Curvature Spectrum (GCS), which preserve global topology and local geometric continuity. The serialized sequences are then processed by Mamba blocks under the Hierarchical Domain-Aware Modeling (HDM) mechanism, which stabilizes intra-domain structure and fuses inter-domain relations. (b) At test time, the Spectral Graph Alignment (SGA) performs structure-aware shifting that guides target features toward source prototypes in the spectral domain without updating model parameters, enabling robust generalization to unseen target domains across multiple tasks.}
    \label{fig:framework}
\end{figure*}

\section{Related Work}
\noindent \textbf{Point Cloud Understanding}. Pioneered by PointNet~\cite{qi2017pointnet} and PointNet++~\cite{qi2017pointnet++}, point-based methods~\cite{lin2023meta,li2024pointgl,lai2022stratified,choe2022pointmixer,wu2024point,ma2022rethinking,li2018pointcnn,wu2019pointconv,wang2019dynamic,thomas2019kpconv,xu2021paconv,komarichev2019cnn,jiang2023masked,jiang2024pcotta} directly learned permutation-invariant features from unordered point sets, while voxel-based~\cite{maturana2015voxnet,graham20183d,qi2016volumetric,choy20194d,tang2020searching}, graph-based~\cite{wang2019dynamic,shi2020point,jiang2024dhgcn}, and projection-based~\cite{wu2018squeezeseg,wu2019squeezesegv2,milioto2019rangenet++,zhang2020polarnet,cortinhal2020salsanext,ando2023rangevit,li2024rapid} captured local geometry. Transformer-based models~\cite{guo2021pct,zhao2021pointtransformer,pang2022masked,lai2022stratified,ando2023rangevit,wu2022point,wu2024point,li2025dapointr} achieved strong global reasoning but suffer from quadratic complexity and weak structural continuity. Recent Mamba-based models~\cite{liang2024pointmamba,zhang2025point,han2024mamba3d,bahri2025spectral,zhang2024voxel,lu2025exploring,li2026dapointmamba} enable efficient sequence modeling, but depend on coordinate-based serialization, making them sensitive to rotations and incomplete regions. Besides, they are designed for single-task learning and struggle to generalize to unseen domains.

\noindent \textbf{Point Cloud Domain Generalization} (DG) seeks models that perform well on unseen domains without target data~\cite{yang2025pointdgmamba,long2025diverse,long2024rethinking,long2025domain,zhou2024test,zhou2023instance}. Early DG works emphasize adversarial~\cite{qin2019pointdan,xu2024push}, contrastive~\cite{wei2022learning,xiao20233d,wei2024multi,liu2025rotation} or augmentation-based~\cite{lehner20223d,xiao2022learning,zhao2024unimix,he2025domain,kim2024rethinking} and consistency-based alignment~\cite{kim2023single,he2025domain,park2025no}, but neglect multi-task learning. Mamba-based DG frameworks~\cite{yang2025pointdgmamba,long2024dgmamba} improve efficiency but still rely on coordinate serialization. Recently, DG-PIC~\cite{jiang2024dg} is the first work that unifies multiple 3D tasks via Transformer prompts but remains computationally expensive and order-agnostic. Nevertheless, all these methods neglect learning the structure-preserving representations that remain invariant to domain and task variations. 

\noindent \textbf{Structure Modeling in Point Clouds} has recently been explored to preserve the structural hierarchy of point clouds. Some researchers design structure-aware operators~\cite{tchapmi2019topnet,he2020structure,wang2020structure,cheng2023structure,li2018so,yang2026pointdgrwkv} to encode local topology or surface continuity. Others construct graph-based representations~\cite{hao2022structure,wang2019graph,liu2023grab,wei2020view,zhang2018graph,hu2020feature,thanou2016graph,shen2018mining} to capture neighborhood relations. In addition, studies~\cite{montanaro2022rethinking,bi2025dual,feng2025hyperbolic,montanaro2023towards,onghena2023rotation,sur2025hyperbolic,li2025hyperbolic,xie2024hecpg} further explore hyperbolic geometry to embed hierarchical or tree-like structures in non-Euclidean space. However, these methods focus on single-domain or single-task structure modeling, but overlook the shared inherent structure across domains and tasks. In contrast, our approach is the first \emph{structure-aware domain generalization} framework for multi-task point cloud understanding, explicitly preserving global topology and local geometry under domain shifts.

%% file: sections/3_method.tex
\section{Methodology}

\subsection{Problem Setting and Overview}
We study domain generalization for point cloud understanding in a multi-domain, multi-task setting, following DG-PIC~\cite{jiang2024dg}. Let $\{D_s^k\}_{k=1}^{K}$ denote $K$ source domains and $D_t$ an unseen target domain. The model is trained on $\{D_s^k\}$ and generalizes to $D_t$ at test time without parameter updates. 
DG-PIC is formulated with In-Context Learning (ICL): given a point cloud, Farthest Point Sampling (FPS) and $k$-Nearest Neighbor (KNN) grouping produce patch tokens $\mathcal{T} = \{t_i\}_{i=1}^N$. A Transformer-based masked autoencoder reconstructs query tokens from prompts, enabling a unified architecture for multiple tasks (\textit{i.e.,} reconstruction, denoising, and registration) under a DG paradigm. 
However, this has drawbacks: (1) quadratic self-attention complexity limits scalability; (2) tokenization lacks explicit ordering, hindering sequential and structural consistency. Consequently, intrinsic geometric and topological cues are not fully captured, weakening generalization under domain shifts.

Motivated by this, we develop a Structure-Aware Domain Generalization (SADG) framework for point clouds. To our knowledge, this is the first work to introduce Mamba into ICL for domain-generalized multi-task point cloud understanding with structural consistency. Figure~\ref{fig:framework} shows that SADG first serializes unordered tokens into transformation-invariant sequences preserving topological and geometric relations, followed by a hierarchical domain-aware mechanism that captures intra-domain structure and inter-domain consistency. In testing, a spectral graph alignment module treats serialized target features as graph signals and aligns them with source prototypes in the spectral domain. These components preserve structural continuity, enhance sequential stability, and improve generalization across domains.

\subsection{Structure-Aware Serialization with Mamba}
\begin{figure}[htbp]
    \centering
    \includegraphics[width=\linewidth]{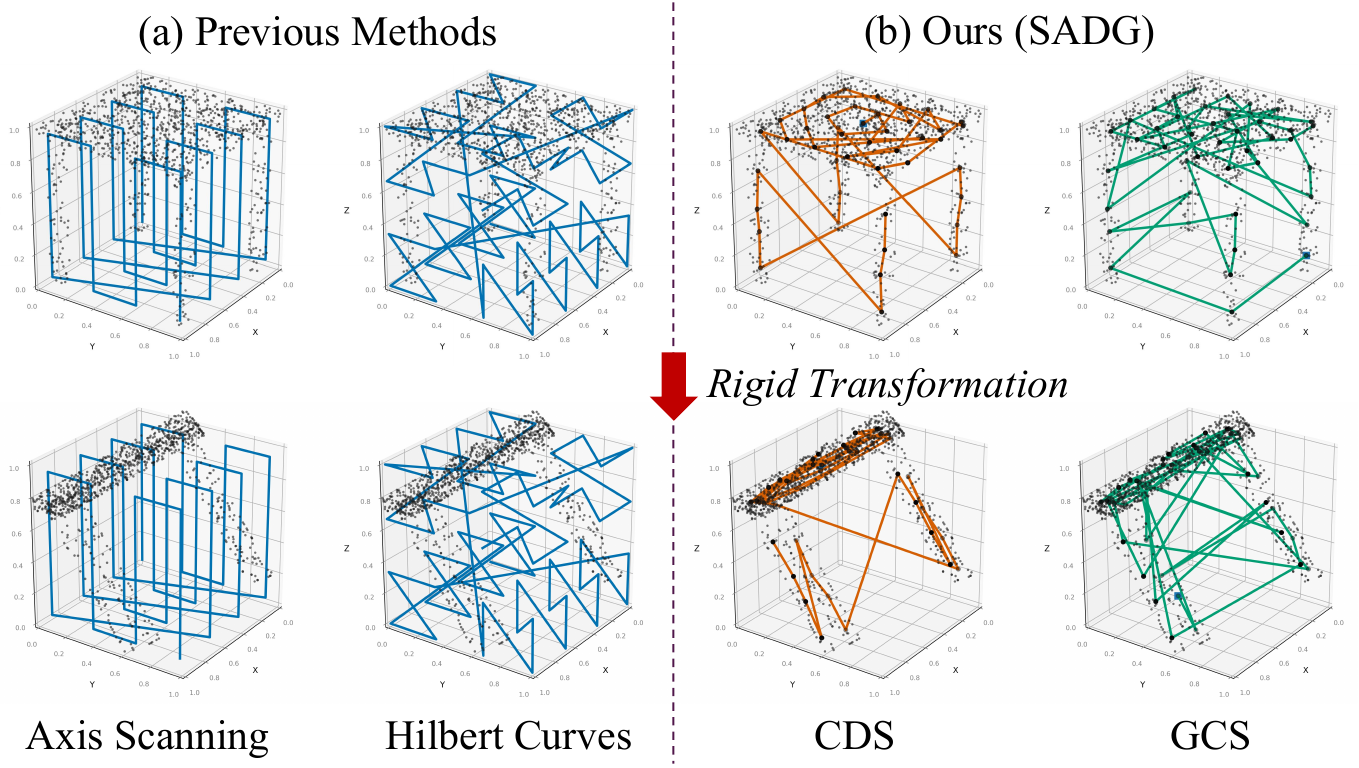}
    \caption{
    Comparison of different serialization strategies. The proposed CDS and GCS maintain transformation invariance and structural consistency across unaligned real-scan objects, providing a stable foundation for domain-generalized 3D understanding.
    }
    \label{fig:teaser}
\end{figure}

To overcome Transformer inefficiency and unordered tokenization, we adopt Mamba~\cite{gu2024mamba} as the sequential backbone for point cloud ICL. However, Figure~\ref{fig:teaser} shows that Mamba is inherently order-sensitive: without a stable sequence, its recurrent updates become fragile and fail to capture structural relationships across domains and tasks. Thus, instead of coordinate-based token ordering, we introduce structure-aware serialization, which constructs intrinsic graph-based sequences encoding topological layout and geometric continuity, providing efficient and consistent inputs for Mamba.

\noindent\textbf{Notation.}
Following DG-PIC~\cite{jiang2024dg}, FPS and KNN grouping produce $N$ local patches (tokens) $\mathcal{T}=\{t_i\}_{i=1}^{N}$ with centers $u_i\in\mathbb{R}^3$ and features $x_i\in\mathbb{R}^{d}$. We construct a token graph 
$\mathcal{G}=(\mathcal{V},\mathcal{E},w)$, where $\mathcal{V}=\{1,\dots,N\}$ indexes tokens and $w(i,j)$ denotes the affinity between tokens $t_i$ and $t_j$. A serialization is defined as a permutation:
\begin{equation}
\pi:\{\mathcal{V}_{1},\ldots,\mathcal{V}_{N}\}\xrightarrow{\text{serialize}}\{\mathcal{V}_{\pi(1)},\ldots,\mathcal{V}_{\pi(N)}\},
\end{equation}
which reorders tokens into a sequence:
\begin{equation}
X_{\pi} = [x_{\pi(1)}, \ldots, x_{\pi(N)}],
\end{equation}
and is then processed by Mamba:
\begin{equation}
Z = \mathrm{Mamba}(X_{\pi}) = [z_{\pi(1)},\ldots,z_{\pi(N)}].
\end{equation}
Different choices of $w(i,j)$ yield different structure-aware serialization spectra, which we will introduce below.

\noindent\textbf{Centroid Distance Spectrum (CDS).} 
To model the structural layout of unordered tokens, we establish a topology-aware serialization based on intrinsic spatial relationships. Given token centers $\{u_i\}_{i=1}^{N}$, we compute the point cloud-wise global centroid: $c = \frac{1}{N}\sum_{i=1}^{N}u_i.$
A naive choice would sort tokens directly by their distances to $c$, \ie, $d_i = \|u_i - c\|_2$, which indeed provides a global measure but neglects local spatial continuity. Such sorting often causes abrupt transitions between spatially distant tokens, disrupting the topological smoothness essential for stable sequential modeling (see Figure~\ref {fig:compare_cds_gcs} in Ablation). To preserve both local continuity and global coverage, we construct a token graph $\mathcal{G}_{CDS} = (\mathcal{V}, \mathcal{E}, w_{CDS})$ with affinity:
\begin{equation}
\label{eq:cds}
w_{CDS}(i,j) = \exp\!\Big(-\frac{\|u_i - u_j\|_2^2}{\sigma^2}\Big),
\end{equation}
which softly connects nearby tokens and suppresses remote ones, thereby preserving local geometric coherence during traversal. Starting from the centroid-nearest token $t_r=\arg\min_i\|u_i-c\|_2$, we perform a Breadth-First Search (BFS) over $\mathcal{G}_{CDS}$ to establish the serialization order. At each step, the current node expands to unvisited neighbors ranked by $w_{CDS}(i,j)$, ensuring that spatially adjacent tokens are explored in a locally smooth manner. This traversal continues until all tokens are visited, yielding a topology-consistent permutation $\pi_{CDS}$ and the corresponding ordered sequence $X_{\pi_{CDS}}$. 
This serialization balances global coverage and local continuity, forming a coherent sequence encoding coarse-to-fine topological information of the point cloud for stable sequential modeling within Mamba.

\begin{figure*}[htbp]
    \centering
    \includegraphics[width=\textwidth]{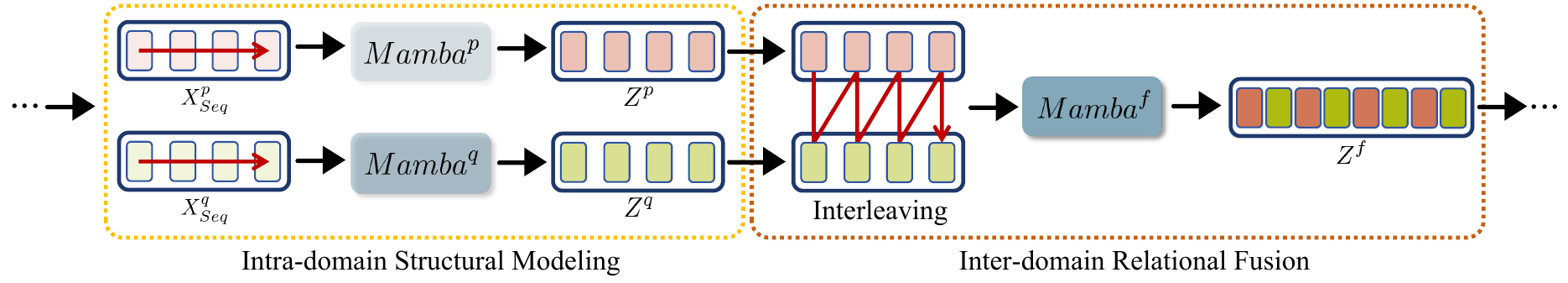}
    \caption{Hierarchical Domain-Aware Modeling (HDM) cascades intra-domain structural modeling and inter-domain relational fusion. 
    }
    \label{fig:hdm}
\end{figure*}

\noindent\textbf{Geodesic Curvature Spectrum (GCS).}
Beyond topology captured by CDS, GCS aims to encode intrinsic surface geometry through curvature-guided diffusion in a geodesic graph. 
General explicit curvature estimation relies on normals or dense sampling, which are fragile under noise, missing regions, and the domain gap between synthetic and real scans (as shown in Figure~\ref {fig:compare_cds_gcs}). 
To overcome this, we formulate curvature implicitly through a heat diffusion process on the geodesic graph, providing a stable and intrinsic representation of local surface geometry.

Since Euclidean distances fail to reflect the intrinsic continuity of curved surfaces, we compute geodesic distances between tokens $t_i$ and $t_j$ as the shortest paths along a local KNN adjacency graph on token centers $\{u_i\}_{i=1}^{N}$:
\begin{equation}
d_{\text{geo}}(i,j) = \min_{\mathcal{P}_{ij}} \sum_{(p,q)\in\mathcal{P}_{ij}}\|u_p - u_q\|_2,
\end{equation}
where $\mathcal{P}_{ij}$ denotes the shortest valid path connecting $t_i$ and $t_j$. 
This formulation follows manifold connectivity and preserves surface-aware consistency across complex geometric regions. 
To this end, we further define a curvature-guided heat diffusion using the Laplace--Beltrami operator $\Delta$ on this geodesic graph. The diffusion equation $\frac{\partial h(t)}{\partial t} = -\Delta h(t)$ implicitly captures local curvature behavior, where highly curved regions dissipate heat faster while flatter regions retain heat longer.
The corresponding heat kernel between tokens $t_i$ and $t_j$ is expressed as:
\begin{equation}
K_\tau(i,j) = \sum_{k=1}^{N} e^{-\lambda_k \tau}\,\phi_k(i)\phi_k(j),
\end{equation}
where $\{\lambda_k,\phi_k\}$ are eigenvalues and eigenfunctions of $\Delta$. 
The diagonal term $K_\tau(i,i)$ measures self-diffusion at node $i$, intrinsically encoding local curvature through diffusion dynamics. 
Sampling across multiple diffusion scales $\{\tau_s\}_{s=1}^{S}$ yields a multi-scale curvature descriptor:
\begin{equation}
h_i = [K_{\tau_1}(i,i), K_{\tau_2}(i,i), \ldots, K_{\tau_S}(i,i)].
\end{equation}
We then define curvature-based affinity between tokens as:
\begin{equation}
\label{eq:gcs}
w_{\text{GCS}}(i,j) = \exp\!\Big(-\frac{\|h_i - h_j\|_2^2}{\gamma^2}\Big),
\end{equation}
and construct the token graph $\mathcal{G}_{\text{GCS}} = (\mathcal{V}, \mathcal{E}, w_{\text{GCS}})$.
Serialization begins from the lowest-curvature token $t_r=\arg\min_i\|h_i\|_2$ and proceeds in ascending curvature order, yielding a permutation $\pi_{\text{GCS}}$ and the corresponding sequence $X_{\pi_{\text{GCS}}}$, which preserves geometric smoothness and curvature coherence across neighboring patches.
This diffusion-based formulation encodes curvature intrinsically via heat propagation, providing stable geometric cues for low-quality data that strengthen Mamba's modeling. 
Together, CDS and GCS provide a transformation-invariant, structure-aware serialization that preserves both topological layout and geometric continuity for sequential modeling.

\noindent\textbf{Unified Structure-Aware Sequence.} 
To enhance contextual modeling with linear-time efficiency, we perform bidirectional traversals on both spectra and concatenate:
\begin{equation}
X_{\text{seq}}=[\,X_{\pi_{\mathrm{CDS}}};X_{\mathrm{rev}(\pi_{\mathrm{CDS}})};X_{\pi_{\mathrm{GCS}}};X_{\mathrm{rev}(\pi_{\mathrm{GCS}})}\,].
\end{equation}
This unified, structure-aware sequence expands Mamba's receptive field while maintaining topological and geometric continuity, allowing Mamba to exploit ordered dependencies without sacrificing efficiency. 

\subsection{Hierarchical Domain-Aware Modeling}
Mamba processes tokens along the serialized order, making it well-suited to preserving topological and geometric continuity. Given the serialized input $X_{\pi}=[x_{\pi(1)},...,x_{\pi(N)}]$, Mamba updates hidden states recurrently:
\begin{equation}
z_t=\mathrm{Mamba}(x_{\pi(t)},z_{t-1})=g(Az_{t-1}+Bx_{\pi(t)}+b),
\end{equation}
where $A,B$ are learnable transition matrices, $b$ is bias, and $g(\cdot)$ the gating function. This update enables linear-time modeling of local continuity and long-range structure.

However, unlike Transformer-based ICL~\cite{jiang2024dg} which uses prompt-query concatenation along the unordered tokens, Mamba is order-sensitive. In DG, the simple concatenation of tokens from different domains disrupts the sequential dynamics and weakens state propagation, leading to unstable cross-domain reasoning. 

To address this, we design a \textbf{Hierarchical Domain-Aware Modeling (HDM)} mechanism that reorganizes serialized features in two cascading stages, as illustrated in Figure \ref{fig:hdm}, enhancing both intra-domain structural modeling and inter-domain relational generalization.

\noindent\textbf{Intra-domain Structural Modeling (ISM).}
Given serialized sequences from prompt and query domains $\{X_{\text{seq}}^{p}, X_{\text{seq}}^{q}\}$, we perform intra-domain modeling to preserve structural dependencies by processing two parallel domain-specific Mamba branches independently:
\begin{equation}
Z^{p}=\mathrm{Mamba}^{p}(X_{\text{seq}}^{p}), \quad 
Z^{q}=\mathrm{Mamba}^{q}(X_{\text{seq}}^{q}).
\end{equation}
This stage stabilizes intra-domain consistency, not only ensuring that stable topological and geometric patterns aggregate within each domain before any cross-domain interaction, but also preventing sequential discontinuities across domain boundaries.

\noindent\textbf{Inter-domain Relational Fusion (IRF).} 
After obtaining $\{Z^{p},Z^{q}\}$, we perform inter-domain relational fusion to establish transferable correspondences across domains. Different from direct concatenation used in Transformer-based ICL~\cite{jiang2024dg}, we interleave tokens from prompt and query domains following their shared structural order $\pi$:
\begin{equation}
Z^{pq} = [\,z_{\pi(1)}^{p},\,z_{\pi(1)}^{q},\,z_{\pi(2)}^{p},\,z_{\pi(2)}^{q},\,\ldots,\,z_{\pi(4N)}^{p},\,z_{\pi(4N)}^{q}\,],
\end{equation}
producing a unified, structurally aligned sequence subsequently processed by a shared Mamba:
\begin{equation}
Z^{f}=\mathrm{Mamba}^{f}(Z^{pq}),
\end{equation}
which jointly models domain-specific and domain-shared dependencies. The interleaved sequence implicitly exchanges features between domains through recurrent propagation without attention-based matching, enhancing relational generalization and structural consistency while maintaining linear efficiency.


\begin{table*}[t]
  \centering
  \caption{Comparison on the multi-domain and multi-task benchmark. All models are trained on four source domains and directly evaluated on the remaining one. Evaluation metric: Chamfer Distance (CD, $\times 10^{-3}$, lower is better).}
  \label{tab:main_results}
  \resizebox{\textwidth}{!}{
  \begin{tabular}{lc|ccc|ccc|ccc|ccc|ccc}
    \hline
    \multirow{2}{*}{\textbf{Method}} & \multirow{2}{*}{\textbf{Setting}}
    & \multicolumn{3}{c|}{\textbf{ModelNet}}
    & \multicolumn{3}{c|}{\textbf{ShapeNet}}
    & \multicolumn{3}{c|}{\textbf{ScanNet}}
    & \multicolumn{3}{c|}{\textbf{ScanObjectNN}}
    & \multicolumn{3}{c}{\textbf{MP3DObject}} \\
    & 
    & Rec. & Den. & Reg.
    & Rec. & Den. & Reg.
    & Rec. & Den. & Reg.
    & Rec. & Den. & Reg.
    & Rec. & Den. & Reg. \\
    \hline
    PointNet~\cite{qi2017pointnet} & General & 20.56 & 27.15 & 17.19 & 19.84 & 32.99 & 19.57 & 23.73 & 30.27 & 19.49 & 21.74 & 33.97 & 21.62 & 22.63 & 35.24 & 23.17 \\
    DGCNN~\cite{wang2019dynamic} & General & 19.17 & 26.91 & 17.41 & 21.18 & 26.97 & 19.05 & 21.73 & 30.35 & 17.51 & 24.88 & 33.08 & 18.85 & 21.83 & 38.10 & 21.82 \\
    PCT~\cite{guo2021pct} & General & 16.98 & 25.10 & 14.50 & 18.39 & 24.59 & 15.28 & 18.76 & 27.03 & 16.99 & 19.50 & 29.98 & 15.71 & 18.91 & 28.74 & 16.49 \\
    Point-MAE~\cite{pang2022masked} & General & 14.77 & 21.53 & 13.42 & 16.82 & 23.91 & 14.36 & 16.16 & 24.54 & 16.79 & 19.27 & 28.69 & 15.74 & 20.39 & 27.28 & 17.13 \\
    PointMamba~\cite{liang2024pointmamba} & General & 16.47 & 23.13 & 13.65 & 16.33 & 25.24 & 14.96 & 15.61 & 22.43 & 14.30 & 17.16 & 25.44 & 17.64 & 20.16 & 27.08 & 17.40 \\
    \hline
    PointMixup~\cite{chen2020pointmixup} & DG & 17.62 & 29.24 & 16.07 & 18.01 & 26.91 & 17.22 & 18.86 & 30.20 & 19.24 & 21.17 & 30.58 & 21.37 & 22.54 & 30.99 & 18.42\\
    PointCutMix~\cite{zhang2022pointcutmix} & DG & 16.23 & 27.07 & 16.77 & 18.19 & 25.30 & 16.24 & 21.28 & 29.21 & 19.07 & 22.64 & 28.58 & 19.85 & 22.80 & 33.24 & 18.88\\
    PointDGMamba~\cite{yang2025pointdgmamba} & DG & 14.39 & 19.37 & 12.44 & 14.56 & 24.05 & 14.22 & 14.67 & 23.10 & 12.97 & 18.19 & 27.07 & 14.51 & 17.99 & 26.82 & 14.66 \\
    \hline
    PIC~\cite{fang2023explore} & ICL & 17.89 & 25.70 & 16.15 & 17.96 & 24.10 & 15.50 & 16.90 & 30.46 & 18.10 & 21.75 & 29.42 & 16.74 & 22.86 & 34.72 & 17.54 \\
    DG-PIC~\cite{jiang2024dg} & ICL+DG &  6.84 &  9.40 &  5.01 &  8.02 &  9.81 &  7.41 &  5.21 &  9.71 &  5.10 &  4.52 & 12.74 &  4.17 &  5.91 & 10.40 &  5.64 \\
    Vanilla Mamba ICL & ICL+DG &  7.69 & 10.81 &  6.22 &  7.98 & 10.19 &  \textbf{6.25} &  5.45 & 10.75 &  5.56 &  6.93 & 11.52 &  7.76 &  8.28 & 14.19 &  8.44  \\
    \textbf{Ours (SADG)} & ICL+DG & \textbf{5.99} & \textbf{7.98} & \textbf{3.81} &
                         \textbf{7.64} & \textbf{9.34} & 7.06 &
                         \textbf{2.97} & \textbf{7.67} & \textbf{3.63} &
                         \textbf{4.29} & \textbf{9.84} & \textbf{3.03} &
                         \textbf{3.55} & \textbf{6.61} & \textbf{2.84} \\
    \hline
  \end{tabular}
  }
\end{table*}

\subsection{Spectral Graph Alignment}

At test time, the model parameters remain frozen, and the goal is to preserve structural consistency on unseen domains. 
We propose a lightweight \textbf{Spectral Graph Alignment (SGA)} performing structure-aware alignment in the spectral domain before Mamba processing. 
Without requiring weight updates, SGA conducts spectral shifting on the latent graphs from CDS and GCS, ensuring topology- and geometry-consistent representations under domain shifts.


For each serialization strategy $*\in\{\text{CDS},\text{GCS}\}$, let $X_{\pi_*}$ denote the serialized token sequence, which we treat as a graph signal on $\mathcal{G}_*=(\mathcal{V},\mathcal{E},w_*)$. 
Using the normalized Laplacian $\mathbf{L}_*=\mathbf{D}_*-\mathbf{A}_*$, the Graph Fourier Transform (GFT) projects the sequence into the spectral domain as $\hat{X}_{*}=\Phi_{*}^{\top}X_{\pi_*}$, where $\Phi_{*}$ denotes the eigenvectors of $\mathbf{L}_*$ serving as the structural frequency bases. 

To guide the domain generalization, we derive two source prototypes, $\hat{P}_{\text{CDS}}^{s}$ and $\hat{P}_{\text{GCS}}^{s}$, which are computed by averaging source domain features and projecting them onto the unified query-specific spectral basis:
\begin{equation}
\hat{P}_{*}^{s}=(\Phi_{*}^{t})^{\top}\!\left(\frac{1}{N_s}\sum_{i=1}^{N_s}X_{\pi_*,i}^{s}\right).
\end{equation}
These prototypes capture domain-level statistics as stable structural anchors for spectral alignment. During testing, target spectral tokens $\hat{X}_{*}^{t}$ align toward the prototypes:
\begin{equation}
\hat{X}_{*,i}^{t}\!\leftarrow\!\alpha_i\,\hat{X}_{*,i}^{t}+(1-\alpha_i)(\hat{P}_{*}^{s}-\hat{X}_{*,i}^{t}),
\end{equation}
where the adaptive coefficient $\alpha_i$ is modulated by the cosine similarity between $\hat{X}_{*,i}^{t}$ and $\hat{P}_{*}^{s}$, enforcing coherent alignment while avoiding over-correction in irregular regions. 
The aligned spectral features are then transformed back to the spatial domain via the inverse GFT as $X_{\pi_*}^{t}=\Phi_{*}^{t}\hat{X}_{*}^{t}$.

By leveraging the intrinsic graphs of CDS and GCS, this spectral process preserves topological and geometric consistency while explicitly mitigating domain discrepancies in the spatial space. Consequently, SGA performs test-time, structure-aware alignment without weight updates, providing a transformation-invariant foundation that enables the model to better generalize to unseen target domains.

%% file: sections/4_experiments.tex
\section{Experiments}

\subsection{Benchmark and Implementation Details}

\noindent\textbf{Multi-domain Multi-task Benchmark.}
Following DG-PIC~\cite{jiang2024dg}, we evaluate our method under a unified multi-domain, multi-task setting. 
The benchmark integrates five datasets with consistent category definitions across seven shared classes (\textit{chair}, \textit{table}, \textit{sofa}, \textit{bed}, \textit{cabinet}, \textit{shelf}, \textit{monitor}), including two synthetic (\textit{ModelNet40}, \textit{ShapeNet}) and three real-scan domains (\textit{ScanNet}, \textit{ScanObjectNN}, and the newly introduced \textit{MP3DObject}). 
All point clouds are sampled to 1,024 points and normalized within a unit sphere. 
Three representative tasks, \textit{i.e.,} reconstruction, denoising, and registration, are learned jointly within a single unified model. 
This benchmark covers diverse domains and geometric conditions, offering a challenging yet comprehensive testbed for generalization across multiple tasks.

\noindent\textbf{MP3DObject.}
To enable more realistic evaluation beyond DG-PIC~\cite{jiang2024dg}, we construct \textit{MP3DObject} from the large-scale indoor dataset Matterport3D~\cite{Matterport3D} by extracting object-level instances and removing extremely incomplete samples. 
Each object is centered and normalized but not aligned to a canonical orientation, introducing natural viewpoint and pose variations. 
This dataset contains 4,015 training samples and 1,003 testing samples, offering real-world variation while preserving complex geometry, making it a challenging benchmark for realistic point cloud understanding.

\noindent\textbf{Implementation.}
All experiments are implemented in PyTorch with CUDA~11.8 and trained on a TITAN RTX GPU using AdamW with learning rate of $1\times10^{-4}$, cosine decay and batch size of 96 for 300 epochs. 
We set $\sigma$ and $\gamma$ to the medians of corresponding token graphs, yielding a robust and adaptive affinity scale.
Following the leave-one-domain-out protocol, models are trained on four source domains and directly evaluated on the unseen target without parameter updates. 
We use the Chamfer Distance (CD) metric to measure geometric consistency between predictions and ground truth across all tasks.

\subsection{Main Results}
\noindent\textbf{General-purpose Point Cloud Baselines.}
We compare our method with approaches for general point cloud learning, including PointNet~\cite{qi2017pointnet}, DGCNN~\cite{wang2019dynamic}, PCT~\cite{guo2021pct}, Point-MAE~\cite{pang2022masked}, and PointMamba~\cite{liang2024pointmamba}. 
All baselines are retrained under the same multi-domain, multi-task protocol, training on four sources and evaluating on the held-out target. 
While competitive in domain-specific or single-task settings, these models degrade in our unified benchmark (Table~\ref{tab:main_results}), with PointNet dropping sharply on real scans such as MP3DObject.
Even modern backbones like Point-MAE and PointMamba still fail to generalize, \textit{e.g.}, PointMamba reaches $20.16/27.08/17.40$ on MP3DObject, indicating sensitivity to viewpoint changes and incompleteness.  
These results highlight that without structure-aware design or domain modeling, general models fail to generalize across heterogeneous domains.

\begin{table}[t]
  \centering
  \caption{Ablation study on the proposed components.}
  \label{tab:ablation_combined}
  \resizebox{\columnwidth}{!}{
  \begin{tabular}{lccc}
    \hline
    Variant & Reconstruction & Denoising & Registration \\
    \hline
    \multicolumn{4}{l}{\textit{Serialization Variants}} \\
    \hline
    Z-order Scanning & 7.32 & 12.47 & 6.29 \\
    Hilbert Curve & 6.23 & 11.13 & 7.68 \\
    w/o CDS & 4.75 & 10.69 & 6.17 \\
    w/o GCS & 5.82 & 8.92 & 4.43 \\
    \textbf{Ours (full SAS)} & \textbf{3.55} & \textbf{6.61} & \textbf{2.84} \\
    \hline
    \multicolumn{4}{l}{\textit{Hierarchical Domain-Aware Modeling}} \\  
    \hline
    w/o ISM & 5.41 & 11.37 & 7.51 \\
    w/o IRF & 6.92 & 9.12 & 7.57 \\
    \textbf{Ours (full HDM)} & \textbf{3.55} & \textbf{6.61} & \textbf{2.84} \\
    \hline
  \end{tabular}}
\end{table}

\noindent\textbf{Domain Generalization Methods.}
DG approaches enhance robustness by diversifying sources or learning domain-invariant representations.  
PointMixup~\cite{chen2020pointmixup} and PointCutMix~\cite{zhang2022pointcutmix} mix cross-domain samples to encourage feature interpolation, while PointDGMamba~\cite{yang2025pointdgmamba} leverages Mamba for improved domain consistency.  
Although effective under standard DG setups, their performance drops sharply in our multi-domain and multi-task benchmark, where models must simultaneously handle heterogeneous domains and multiple objectives.  
For instance, PointMixup obtains CD errors $18.86/30.20/19.24$ on ScanNet, and PointCutMix reaches $22.80/33.24/18.88$ on MP3DObject, indicating persistent inconsistencies across domains and tasks.  
In contrast, our SADG achieves CD errors $2.97/7.67/3.63$ on ScanNet and $3.55/6.61/2.84$ on MP3DObject, substantially outperforming across all tasks.  
These results show that structure-aware serialization, hierarchical Mamba reasoning, and spectral graph alignment jointly enable coherent topology-geometry modeling across domains, yielding more stable and transferable features under distribution shifts.

\noindent\textbf{In-context Learning Methods.}
We further compare with ICL-based frameworks that unify multiple tasks via prompt conditioning.  
PIC~\cite{fang2023explore} and DG-PIC~\cite{jiang2024dg} demonstrate the benefit of prompt-query design, yet their Transformer backbones and unsorted token sequences hinder efficiency and stability. 
DG-PIC improves performance (\textit{e.g.}, $6.84/9.40/5.01$ CD on ModelNet) but remains constrained by quadratic attention and order-agnostic tokenization.  
To assess backbone influence, we implement a vanilla Mamba-based ICL by replacing DG-PIC's Transformer with Mamba blocks.  
Although more efficient, this variant shows unstable behavior (\textit{e.g.}, $8.28/14.19/8.44$ on MP3DObject) due to coordinate-based tokenization being highly sensitive to pose variation and incompleteness.  
With our Structure-Aware Serialization (SAS) and Hierarchical Domain-Aware Modeling (HDM), both stability and generalization improve markedly.  
As shown in Table~\ref{tab:main_results}, our model achieves $5.99/7.98/3.81$ on ModelNet, $7.64/9.34/7.06$ on ShapeNet, and the best results across all real-world scenarios like MP3DObject.  
These results confirm that SAS enables coherent topology-geometry reasoning in Mamba, while HDM and SGA further ensure robust generalization to unseen targets.

\begin{figure}[tbp]
    \centering
    \includegraphics[width=\linewidth]{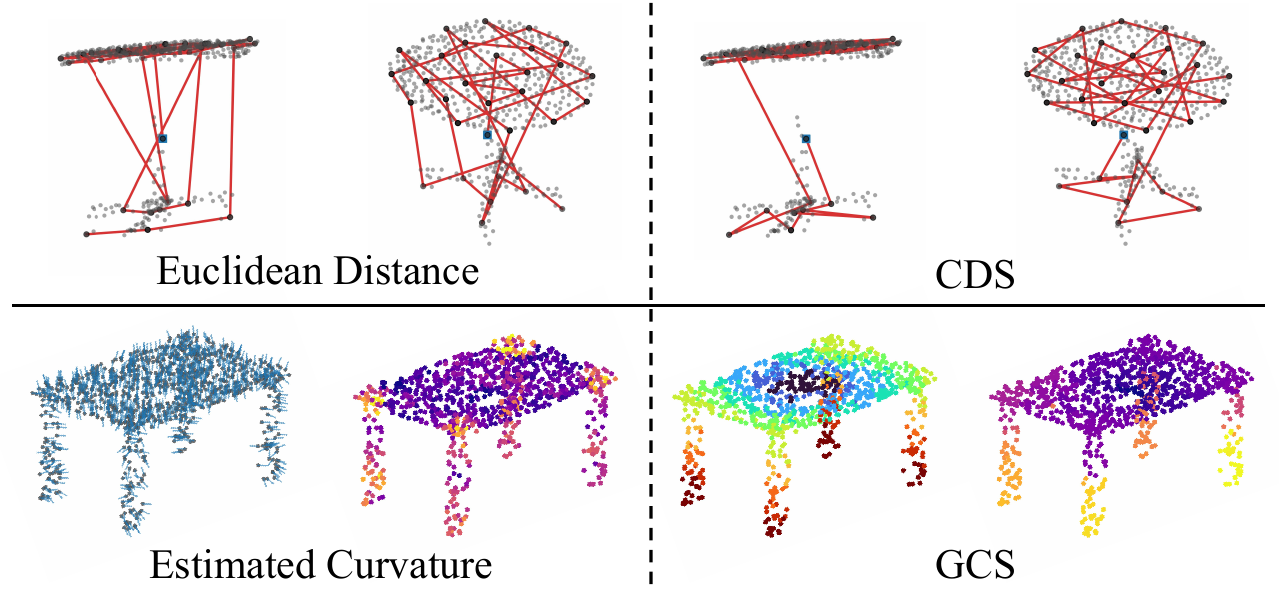}
    \caption{Comparison between naive serialization variants and our method, highlighting stable topology and geometry in 3D objects.}
    \label{fig:compare_cds_gcs}
\end{figure}

\begin{figure*}[htbp]
    \centering
    \includegraphics[width=0.99\linewidth]{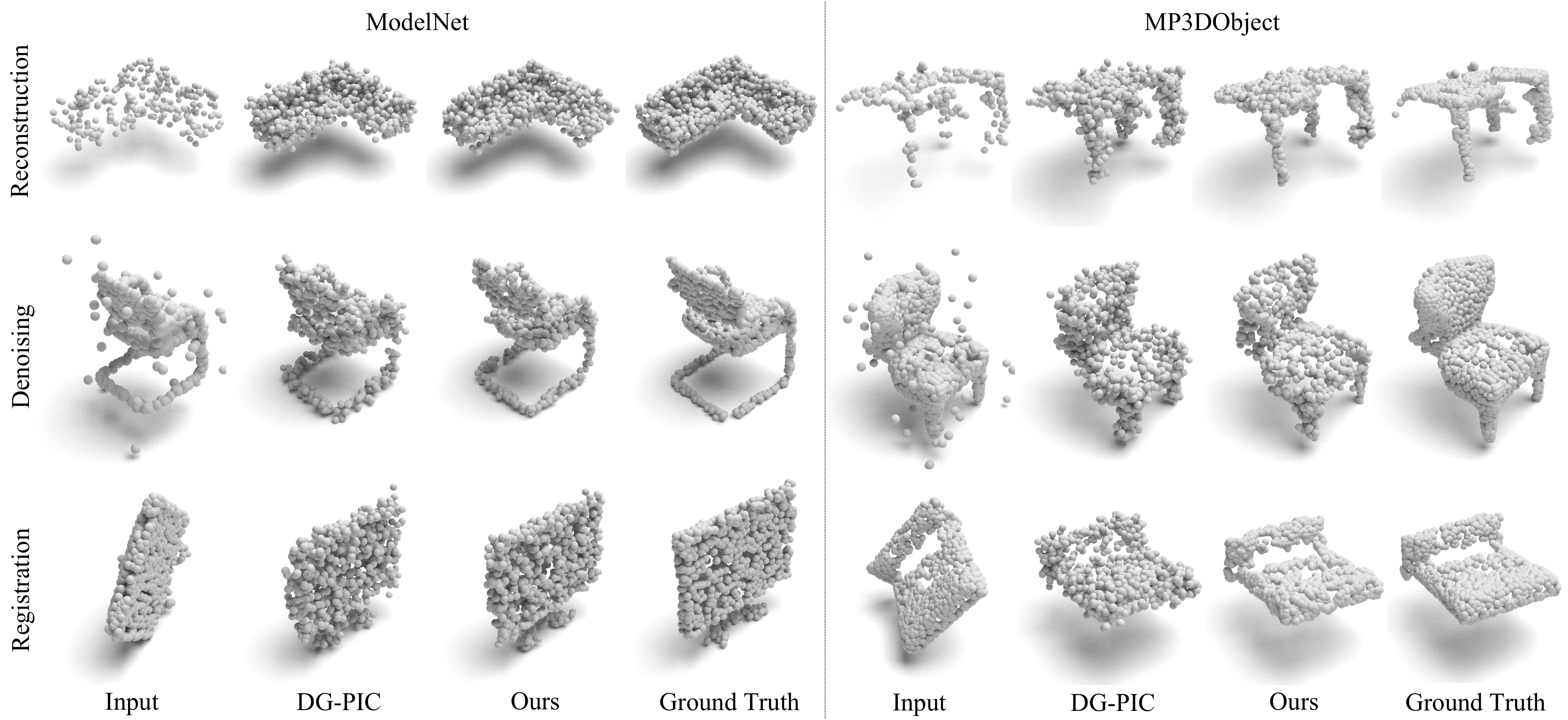}
    \caption{Qualitative results on synthetic (ModelNet) and real-scan (MP3DObject) targets. Our method recovers both detailed geometry and smooth surfaces. For visualization, MP3DObject instances are rendered after aligning their non-canonical orientations. }
    \label{fig:qualitative}
\end{figure*}

\subsection{Ablation Studies}
We analyze each component using MP3DObject as targets, which contains complex geometries and unaligned poses. 

\noindent\textbf{Serialization Variants.}
We compare several serialization strategies to validate our structure-aware design.  
As shown in Table~\ref{tab:ablation_combined}, coordinate-based traversals such as Z-order and Hilbert yield higher CD errors (\textit{e.g.,} $7.32/12.47/6.29$ and $6.23/11.13/7.68$), reflecting their sensitivity to orientation and lack of intrinsic structure embedding.  
Removing CDS or GCS further degrades performance ($4.75/10.69/6.17$ and $5.82/8.92/4.43$), 
confirming that topology and geometry provide complementary cues for stable sequential modeling.  
Figure~\ref{fig:compare_cds_gcs} illustrates the naive options for our serialization. 
Top-left shows a simple centroid Euclidean distance ordering where the sequence repeatedly jumps across the surface, breaking local continuity.
Bottom-left depicts curvature-based sorting, whose estimates are noisy on real scans and produce irregular surface ordering.
Our design (\textit{i.e.,} CDS and GCS) stably models topology and geometry relations, maintains strong structural coherence and domain robustness across multiple point cloud understanding tasks.

\noindent\textbf{Hierarchical Domain-Aware Modeling.}
HDM performs intra-domain structural modeling and inter-domain relational fusion within unified Mamba sequences.  
Due to Mamba's order sensitivity, direct concatenation of diverse domains disrupts state transitions.  
Table~\ref{tab:ablation_combined} shows that removing ISM increases CD errors to $5.41/11.37/7.51$, while removing IRF yields $6.92/9.12/7.57$, both noticeably worse than our full HDM ($3.55/6.61/2.84$).  
These results verify that HDM stabilizes sequential updates and preserves consistent structural reasoning across domains.

\begin{figure}[htbp]
    \centering
    \includegraphics[width=\linewidth]{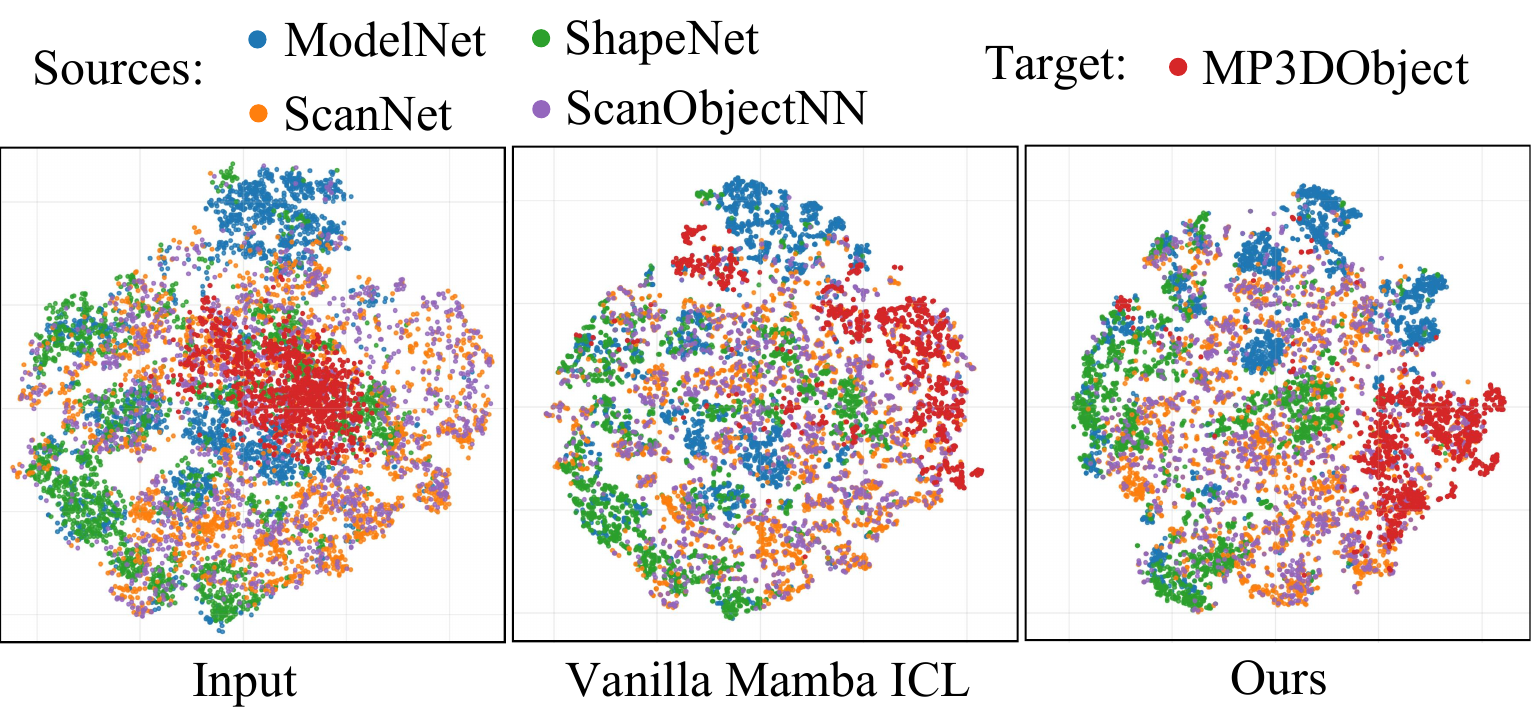}
    \caption{t-SNE visualization of latent features. The Input plot shows encoder-only features, while the latter two depict representations after Mamba-based sequential modeling.}
    \label{fig:tsne}
\end{figure}

\subsection{Visualization and Efficiency Analysis}

\noindent\textbf{Qualitative Results.}
Figure~\ref{fig:qualitative} shows results for reconstruction, denoising, and registration.  
On synthetic targets such as ModelNet, our model restores fine details while preserving overall geometry.  
On real scans like MP3DObject, it achieves higher structural fidelity with clearer boundaries, fewer holes, and smoother surfaces.  
Notably, our method also better preserves thin structures and recovers missing surface regions.  
These qualitative improvements confirm that SAS provides robust topology and geometry cues across domains, particularly under noise and incompleteness.  
Moreover, the consistent reconstructed shapes across synthetic and real data indicates that our SADG framework maintains coherent structural reasoning even under severe viewpoint shifts and partial observations, further highlighting its advantage over existing baselines.  
More qualitative examples are provided in the supplementary material.

\noindent\textbf{t-SNE Visualization.}
To evaluate domain alignment, we visualize Mamba features using t-SNE in Figure~\ref{fig:tsne}.
Vanilla Mamba ICL produces largely domain-separated clusters, indicating weak transferability.  
Our method enables compact intra-domain clusters and stronger overlap between source and unseen targets, showing improved cross-domain alignment without collapsing domain-specific structure, yielding more stable and generalizable embeddings with reduced feature fragmentation under severe distribution shifts, thus strengthening domain generalization.

\noindent\textbf{Efficiency Analysis.}
We compare runtime and model complexity with DG-PIC~\cite{jiang2024dg}.  
Using the same input, we achieve 0.75s inference with 18.87M parameters and 14.89G FLOPs, versus DG-PIC's 0.94s, 27.57M, and 21.07G FLOPs, while delivering better results.  
This highlights the efficiency and scalability of our Mamba-based design, offering an improved efficiency-performance trade-off under the multi-domain and multi-task setting.

%% file: sections/5_conclusion.tex
\section{Conclusion}
We introduced a structure-aware domain generalization framework for multi-task point cloud understanding, which incorporates intrinsic topology and geometry into sequence modeling. 
By constructing the Centroid Distance Spectrum and Geodesic Curvature Spectrum, unordered tokens are serialized into transformation-invariant, structure-consistent sequences, enabling Mamba to model long-range dependencies efficiently and stably. 
To enhance cross-domain generalization, the proposed Hierarchical Domain-Aware Modeling performs intra-domain structural reasoning and inter-domain relational fusion within a unified sequential representation, while the Spectral Graph Alignment ensures structure-aware feature alignment at test time without parameter updates.
Extensive experiments verify that our SADG achieves superior performance on various datasets. 

%% file: sections/X_suppl.tex
\clearpage
\setcounter{page}{1}
\maketitlesupplementary


This document provides additional technical details, ablation studies, and qualitative evaluations to complement the main paper. The contents are organized as follows:
\begin{itemize}
  \item \textbf{A.} Further discussion of our Structure-Aware Serialization (SAS), including the spectral formulation of the Centroid Distance Spectrum (CDS), the motivation for geodesic graphs in the Geodesic Curvature Spectrum (GCS), and empirical evidence of structural drift under domain shifts.
  \item \textbf{B.} More ablation studies on key designs, including serialization strategies, Hierarchical Domain-Aware Modeling (HDM), Spectral Graph Alignment (SGA), and the complexity/runtime breakdown of the full framework. Unless noted, all ablations use MP3DObject as the target domain and report Chamfer Distance (CD) on reconstruction, denoising, and registration.
  \item \textbf{C.} Details of the MP3DObject dataset, including construction, qualitative comparisons with existing benchmarks, and class-wise visualizations in original versus aligned poses.
  \item \textbf{D.} Additional qualitative comparisons, with separate figures for all different target domains and tasks.
  \item \textbf{E.} Training and architectural hyperparameters corresponding to the released codes.
\end{itemize}

\section*{A. Structure-Aware Serialization: Further Discussion}

The main paper introduces Structure-Aware Serialization (SAS) composed of the Centroid Distance Spectrum (CDS) and the Geodesic Curvature Spectrum (GCS), which together produce transformation-invariant and structure-consistent sequences for Mamba. Here we provide additional explanation of the CDS implementation and the choice of geodesic graphs for GCS.

\subsection*{A.1. Centroid Distance Spectrum (CDS)}
\noindent\textbf{GPU-friendly Spectral formulation of CDS.}
CDS is designed to impose a topology-aware ordering over patch tokens by expanding from the centroid along intrinsic surface connectivity. A literal implementation would first build a KNN graph over token centers and then perform a breadth-first search (BFS) from the centroid-nearest node. Although conceptually simple, this queue-based traversal is inherently sequential: each frontier must be fully expanded before the next level, leading to poor utilization of GPU parallelism during large-batch training.

To obtain a topology-coherent ordering without relying on sequential queue operations, we adopt a GPU-compatible spectral formulation. Based on the token graph $\mathcal{G}_{\mathrm{CDS}}$, we compute its normalized graph Laplacian~$\mathbf{L}$. Let $\phi_1$ denote the first non-trivial eigenvector of the generalized eigenproblem $\mathbf{L}\phi_k=\lambda_k\phi_k$. CDS then orders tokens by sorting their scalar embeddings:
\begin{equation}
\pi_{\mathrm{CDS}}(i) = \operatorname{argsort}\!\big(\phi_1(i)\big).
\end{equation}
The Fiedler vector varies smoothly over well-connected regions, assigning nearby scalar values to intrinsically adjacent tokens. Consequently, the resulting serialization follows a topology-aware progression that stimulates the intended BFS behavior, while being fully realizable via batched linear-algebra operations on GPUs.

\noindent\textbf{Neighborhood preservation analysis.}
To quantify how faithfully spectral CDS reflects the BFS expansion pattern, we introduce a neighborhood preservation rate (NPR) with naive BFS as reference ordering. For token $i$, let $\mathcal{N}^{\text{BFS}}_r(i)$ and $\mathcal{N}^{\text{CDS}}_r(i)$ denote its $r$-hop neighborhoods under naive BFS and spectral CDS, respectively, both defined on the same KNN graph. NPR is computed as
\begin{equation}
\mathrm{NPR} = \frac{1}{G}\sum_{i=1}^{G}
\frac{\left|\mathcal{N}^{\text{BFS}}_r(i)\cap \mathcal{N}^{\text{CDS}}_r(i)\right|}
{\left|\mathcal{N}^{\text{BFS}}_r(i)\right|},
\end{equation}
where $G$ is the number of tokens. NPR captures how well the spectral ordering preserves local neighborhoods induced by BFS  and serves purely as an auxiliary diagnostic.

\noindent\textbf{Comparison with naive BFS.}
We implement a naive BFS on CPU, computing the exact BFS layer sequence for reference. We then compare naive BFS and spectral CDS in terms of their neighborhood preservation rate (NPR), average Chamfer Distance (CD) on MP3DObject under reconstruction, denoising, and registration, and per-batch runtime, as shown in Table~\ref{tab:supp_bfs_vs_spectral}. NPR is computed with $r=2$ hops by default, providing a locality-sensitive measure of how well the spectral ordering preserves BFS neighborhoods.

\begin{table}[htbp]
\centering
\caption{Naive BFS vs.\ spectral CDS under MP3DObject as target.}
\label{tab:supp_bfs_vs_spectral}
\resizebox{\columnwidth}{!}{
\begin{tabular}{lccc}
\toprule
Variant & NPR $\uparrow$ & Avg.\ CD $\downarrow$ & Runtime / batch $\downarrow$ \\
\midrule
Naive BFS (CPU)          & 1.00 & 4.27 & 2.22s \\
Spectral CDS (GPU, ours) & 0.97 & 4.33 & 0.75s  \\
\bottomrule
\end{tabular}}
\end{table}

\noindent\textbf{Practical usage.}
Naive BFS is used only as an offline diagnostic to validate the spectral approximation on a small subset. All training and inference in the main paper employ the GPU-based spectral CDS implementation, which is stable, fully batched, and scalable.

\subsection*{A.2. Geodesic Curvature Spectrum (GCS)}

\begin{figure}[htbp]
  \centering
  \includegraphics[width=\linewidth]{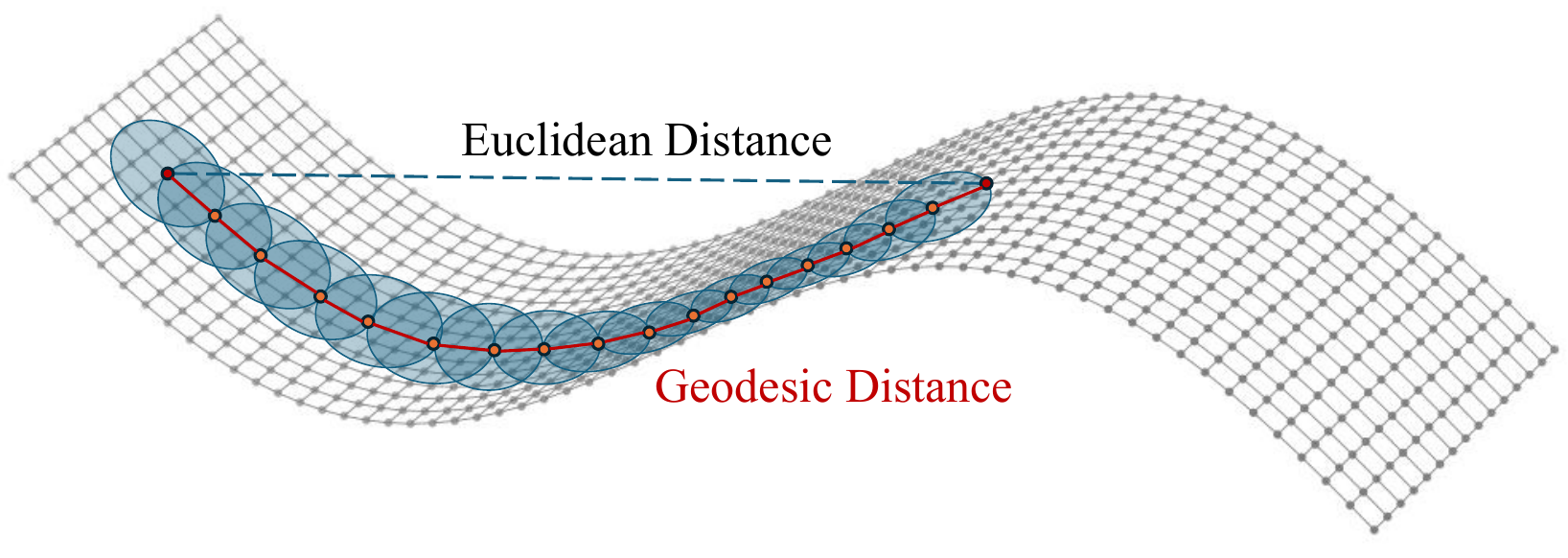}
  \caption{Illustration of why Euclidean distance fails on curved surfaces: straight-line distances shortcut across the token, whereas geodesic distances correctly follow the intrinsic surface geometry.}
  \label{fig:supp_gcs}
\end{figure}

\noindent\textbf{Why Euclidean distances are insufficient.}
GCS is designed to capture local curvature and surface continuity within each patch. A naive approach would be to directly measure Euclidean distances between patch centers and use them as pairwise distances for subsequent diffusion. However, on curved surfaces, Euclidean distances in the ambient space can be misleading: two tokens on opposite sides of a folded surface may be close in straight-line Euclidean distance while being far apart along the surface itself. As illustrated in Figure~\ref{fig:supp_gcs}, a straight segment connecting two tokens across a fold cuts through the volume and ignores curvature, whereas the intrinsic distance along the surface follows the bend.

For heat diffusion and curvature-sensitive descriptors, what matters is connectivity along the surface, not through the volume. Using Euclidean distances directly for diffusion would therefore create artificial shortcuts that bypass folds and holes, breaking the intrinsic continuity of the object.

\noindent\textbf{Geodesic graphs over patch centers.}
To reflect intrinsic surface geometry rather than raw Euclidean proximity, we build a geodesic graph whose nodes correspond to patch centers $\{u_i\}$. Each node is connected to its nearest neighboring patches, producing a locally coherent connectivity that follows the surface layout. Although these connections are established using distances in the ambient space, the resulting graph encourages shortest paths to propagate along the surface, which provides a closer approximation to intrinsic geodesic relations than direct center-to-center measurements.

Based on this graph, we apply heat diffusion at multiple time scales. Short diffusion times emphasize very local geometric variation, while longer times aggregate information over broader neighborhoods of the surface. For each patch, the diffusion responses are consolidated into an intrinsic scalar score that varies smoothly along regions of similar curvature and changes more distinctly across folds or high-curvature transitions. Sorting patches according to this score produces the GCS ordering, which reflects the intrinsic structure of the surface rather than depending on ambient-space distances between patch centers.

\noindent\textbf{Robustness on real-scan domains.}
This graph-based formulation is particularly important on real-scan domains such as ScanObjectNN and MP3DObject, where occlusions, missing regions, and sensor noise are common. In such cases, Euclidean distances between sparse, noisy patch centers can be unstable. In contrast, GCS relies on local connectivity among patches and multi-scale diffusion, which remain more stable under partial observations and provide a consistent, domain-agnostic curvature cue for Mamba.

\subsection*{A.3. Structural Drift Evidence}

\noindent\textbf{Definition of structural drift.}
In the main paper, we show that coordinate-driven serializations are fragile under domain shifts. By \emph{structural drift}, we mean that perturbations such as noise, occlusion, and pose variation can distort sequence-local neighborhoods induced by coordinate-based orders, causing them to deviate from the intrinsic topological or geometric neighborhoods of the underlying shape. This mismatch is particularly harmful to sequence models such as Mamba, whose recurrent updates rely on stable local ordering.

\noindent\textbf{Neighborhood preservation rate under domain shifts.}
Similar to the NPR in Sec.~A.1, we quantify this effect by comparing sequence-local windows with intrinsic neighborhoods. Given a serialization order $\pi$, let $\mathcal{W}^{h}_{\pi}(i)$ denote the local window centered at token $i$ with radius $h$ in the serialized sequence. We then define two intrinsic references:
(1) \emph{Topo-NPR}, where $\mathcal{N}^{\mathrm{topo}}_k(i)$ is the $k$-NN neighborhood of token $i$ on the token graph;
(2) \emph{Geo-NPR}, where $\mathcal{N}^{\mathrm{geo}}_k(i)$ is the top-$k$ neighborhood defined by encoder-feature similarity.
For either definition, the NPR is computed as
\begin{equation}
\mathrm{NPR}(h)=\frac{1}{G}\sum_{i=1}^{G}
\frac{\left|\mathcal{N}_{k}(i)\cap \mathcal{W}^{h}_{\pi}(i)\right|}
{\left|\mathcal{N}_{k}(i)\right|}.
\end{equation}
A higher value indicates that the serialization better preserves intrinsic neighborhoods under domain perturbation.

\noindent\textbf{Empirical comparison under pose perturbation.}
We evaluate the NPR under random rotations as a controlled form of domain shift and compare our CDS/GCS-based serialization against coordinate-driven baselines. As shown in Table~\ref{tab:supp_drift_npr}, Z-order and Hilbert curves yield noticeably lower Topo-NPR and Geo-NPR, indicating stronger structural drift. In contrast, our structure-aware serialization preserves substantially more intrinsic neighborhoods, supporting our claim that SAS reduces drift by maintaining sequence-local neighborhoods that remain more faithful to the underlying geometry.

\begin{table}[htbp]
\centering
\caption{Neighborhood preservation rate under random rotations (higher is better). Our serialization preserves intrinsic neighborhoods more faithfully than coordinate-driven baselines.}
\label{tab:supp_drift_npr}
\resizebox{0.8\columnwidth}{!}{
\begin{tabular}{lcc}
\toprule
Method & Topo-NPR $\uparrow$ & Geo-NPR $\uparrow$ \\
\midrule
Z-order & 0.59 & 0.48 \\
Hilbert curve & 0.62 & 0.48 \\
SAS (CDS+GCS, ours) & \textbf{0.69} & \textbf{0.67} \\
\bottomrule
\end{tabular}}
\end{table}

\section*{B. More Ablation Studies}

We present more ablation studies on the main components of Structure-Aware Serialization (SAS), Hierarchical Domain-Aware Modeling (HDM), and Spectral Graph Alignment (SGA). Unless otherwise noted, all experiments follow the cross-domain protocol in which MP3DObject is treated as an unseen target domain. We report Chamfer Distance (CD $\times 10^{-3}$, lower is better) across reconstruction, denoising, and registration.

\subsection*{B.1. Serialization Strategies}

\noindent\textbf{Naive serialization and random traversal.}
The main paper compares coordinate-based serialization, CDS-only, and GCS-only variants. Here we further examine two additional baselines:

(1) \emph{Naive FPS order}: tokens are ordered according to the Farthest Point Sampling (FPS) index, without any reordering;

(2) \emph{Random traversal}: tokens are randomly permuted with a fixed random seed per instance, simulating a Transformer-style input where the model does not receive explicit structural ordering.

We compare these baselines with our full SAS (CDS+GCS) in Table~\ref{tab:supp_serialization_mp3d}.
Naive FPS order already improves over completely unstructured inputs by preserving some spatial coverage, but it ignores intrinsic topology and curvature and therefore underperforms SAS. Random traversal further disrupts structural continuity, producing the worst CD among the three. These results confirm that SAS provides a meaningful and non-trivial serialization signal for Mamba.

\begin{table}[htbp]
\centering
\caption{Ablations of serialization strategies on MP3DObject. Naive FPS order and random traversal are clearly inferior to our structure-aware serialization.}
\label{tab:supp_serialization_mp3d}
\resizebox{\columnwidth}{!}{
\begin{tabular}{lccc}
\toprule
Variant & Reconstruction & Denoising & Registration \\
\midrule
Naive FPS order          & 7.73 & 13.64 & 7.54 \\
Random traversal         & 8.17 & 12.92 & 7.24 \\
SAS (CDS+GCS, ours)      & \textbf{3.55} & \textbf{6.61} & \textbf{2.84} \\
\bottomrule
\end{tabular}}
\end{table}

\noindent\textbf{Fixed vs.\ data-adaptive kernel scales.}
In CDS and GCS, the Gaussian kernels that define edge weights are controlled by scale parameters $\sigma$ and $\gamma$ (as in eq.~(\ref{eq:cds}) and eq.~(\ref{eq:gcs})). In the main model, both are set in a data-adaptive way based on the median pairwise distances in the corresponding graphs, which automatically adjusts to varying object scale and point density.

To examine the effect of this design, we compare the adaptive setting with fixed kernel scales, which are commonly used in graph-based Gaussian weighting. Concretely, we consider:

(1) \emph{Fixed kernel}: $\sigma, \gamma \in \{0.05, 0.1, 0.2\}$ in the normalized coordinate space;

(2) \emph{Adaptive kernel (ours)}: $\sigma$ and $\gamma$ set to the median of local distances in CDS and GCS graphs, respectively.

As summarized in Table~\ref{tab:supp_sigma_gamma_fixed}, the adaptive median-based scales consistently match or slightly surpass the best fixed choices, while eliminating the need for domain-specific tuning. Their data-driven nature allows the kernel scales to automatically adjust to variations in object size, sampling density, and noise patterns, thus mitigating cross-domain shifts and accommodating data captured from different sensors or scanning conditions.

\begin{table}[htbp]
\centering
\caption{Fixed vs.\ data-adaptive kernel scales on MP3DObject. We report average Chamfer Distance (CD $\times 10^{-3}$) over reconstruction, denoising, and registration. Data-adaptive scales based on median distances provide robust performance without tuning.}
\label{tab:supp_sigma_gamma_fixed}
\resizebox{\columnwidth}{!}{
\begin{tabular}{lcc}
\toprule
Kernel setting & Avg.\ CD & Comment \\
\midrule
Fixed $\sigma=\gamma=0.05$ & 5.76 & too local, sensitive to noise \\
Fixed $\sigma=\gamma=0.10$ & 5.02 & tuned baseline \\
Fixed $\sigma=\gamma=0.20$ & 5.52 & overly smooth, loses detail \\
Adaptive (median-based, ours) & \textbf{4.33} & robust across categories \\
\bottomrule
\end{tabular}}
\end{table}

\subsection*{B.2. Interleaving in Hierarchical Domain-Aware Modeling (HDM)}

HDM first performs intra-domain structural modeling and then fuses prompt and query tokens via a global Mamba operating on an interleaved sequence. To assess the role of interleaving, we compare the proposed design to a simple concatenation variant:
\begin{equation}
Z_{\text{concat}} = [Z^p_{\pi(1)},\dots,Z^p_{\pi(4N)},Z^q_{\pi(1)},\dots,Z^q_{\pi(4N)}],
\end{equation}
where $Z^p$ and $Z^q$ denote prompt and query tokens serialized by SAS.
As shown in Table~\ref{tab:supp_hdm_interleave}, concatenation consistently underperforms the interleaved variant across three tasks on MP3DObject. A hard domain boundary in the concatenated sequence restricts state propagation from prompts to queries, while interleaving enforces fine-grained structural alignment between domains and allows Mamba to exploit prompt information at every step.

\begin{table}[htbp]
\centering
\caption{Interleaving vs.\ concatenation in HDM on MP3DObject. Interleaving tokens from both domains enables more effective cross-domain information flow.}
\label{tab:supp_hdm_interleave}
\resizebox{\columnwidth}{!}{
\begin{tabular}{lccc}
\toprule
Variant & Reconstruction & Denoising & Registration \\
\midrule
w/ concat  & 5.40 & 7.11 & 4.23 \\
w/ interleave (ours)      & \textbf{3.55} & \textbf{6.61} & \textbf{2.84} \\
\bottomrule
\end{tabular}}
\end{table}

\subsection*{B.3. Spectral Graph Alignment (SGA)}

\noindent\textbf{SGA vs.\ simple feature shifting.}
To highlight the role of SGA, we compare it with a simple feature shifting strategy that pushes target features towards source prototypes in the feature space:
\begin{equation}
{X}_{*}^{t}\!\leftarrow\!\beta\,{X}_{*}^{t}+(1-\beta)({P}_{*}^{s}-{X}_{*}^{t}),
\end{equation}
where ${P}_{*}^{s}$ denotes source-domain prototypes and $\beta$ is a fixed scalar (set to $0.5$ by default). This baseline does not use spectral decomposition or frequency-aware mixing.

Table~\ref{tab:supp_sga_vs_shift} shows that simple feature shifting recovers part of the domain gap but remains clearly inferior to SGA, particularly on registration. This indicates that aligning in the spectral domain of CDS/GCS graphs, rather than in raw feature space, is crucial for preserving structural consistency.

\begin{table}[htbp]
\centering
\caption{SGA vs.\ simple feature shifting on MP3DObject. Spectral alignment yields consistently better domain generalization.}
\label{tab:supp_sga_vs_shift}
\resizebox{\columnwidth}{!}{
\begin{tabular}{lccc}
\toprule
Variant & Reconstruction & Denoising & Registration \\
\midrule
No SGA                & 11.95 & 17.38 & 9.43 \\
Simple feature shift  & 7.62 & 12.56 & 7.96 \\
SGA (ours)            & \textbf{3.55} & \textbf{6.61} & \textbf{2.84} \\
\bottomrule
\end{tabular}}
\end{table}

\noindent\textbf{Alignment strength.}
SGA employs adaptive cosine-similarity mixing weights that modulate spectral components according to domain affinity. To assess the influence of alignment magnitude, we replace the adaptive weights with a fixed global coefficient $\alpha \in \{0.0, 0.5, 1.0\}$ applied uniformly to the spectral mixing term. As shown in Table~\ref{tab:supp_alpha}, disabling alignment ($\alpha=0.0$) leads to weaker cross-domain consistency, while very strong alignment ($\alpha=1.0$) risks over-correction. A moderate fixed strength ($\alpha=0.5$) improves stability, but the adaptive cosine-similarity scheme remains the most effective overall, as it naturally adjusts to variations across domains, object geometry, and sensor conditions without requiring per-domain tuning.

\begin{table}[htbp]
\centering
\caption{Effect of alignment strength in SGA on MP3DObject. We report average Chamfer Distance (CD $\times 10^{-3}$) over three tasks.}
\label{tab:supp_alpha}
\resizebox{\columnwidth}{!}{
\begin{tabular}{lcc}
\toprule
Setting & Avg.\ CD & Comment \\
\midrule
$\alpha = 0.0$ (no SGA)          & 12.92 & no alignment applied \\
$\alpha = 0.5$ (fixed)           & 6.77 & improves over no alignment \\
$\alpha = 1.0$ (fixed)           & 10.20 & over-alignment, less stable \\
Adaptive (cosine, ours)       & \textbf{4.33} & similarity-based, best overall \\
\bottomrule
\end{tabular}}
\end{table}

\subsection*{B.4. Complexity and Runtime Breakdown}

\noindent\textbf{Complexity analysis.}
A natural question is whether SAS and SGA introduce substantial overhead beyond the Mamba backbone. In our implementation, both modules operate on a patch-token graph with $G$ nodes after FPS+KNN grouping, rather than on the raw point cloud with $P$ points. In all experiments, $G=64 \ll P=1024$, so the additional cost remains at the token level and is much smaller than operating directly on dense point sets.

Let $G$ denote the token number, $S$ the patch size, $d$ the feature dimension, and $L$ the serialized sequence length. CDS and SGA require spectral decomposition on a $G \times G$ token-graph Laplacian, resulting in complexity $O(G^3)$. GCS computes patch-wise spectra on $G$ local graphs of size $S \times S$, leading to complexity $O(GS^3)$. Therefore, the overall overhead introduced by SAS and SGA is
\begin{equation}
T_{\text{SAS+SGA}} = O(G^3 + GS^3).
\end{equation}
For sequential backbones, Mamba scales linearly as $O(Ld)$, while Transformer self-attention scales quadratically as $O(L^2 d)$.

\noindent\textbf{Runtime breakdown.}
Table~\ref{tab:supp_runtime} reports the runtime breakdown of SADG, including SAS, Mamba forward, and SGA, together with FLOPs and parameter counts. Although SAS and SGA introduce additional graph computations, the total runtime of SADG remains lower than DG-PIC due to the linear-time Mamba backbone. This confirms that our structure-aware design improves efficiency while preserving strong domain generalization performance.

\begin{table}[htbp]
\centering
\caption{Runtime breakdown and model complexity.}
\label{tab:supp_runtime}
\resizebox{\columnwidth}{!}{
\begin{tabular}{lcccccc}
\toprule
Method & SAS (s) & Fwd. (s) & SGA (s) & Total (s) & FLOPs (G) & Params (M) \\
\midrule
DG-PIC~\cite{jiang2024dg} & -- & 0.94 & -- & 0.94 & 21.07 & 27.57 \\
SADG (ours) & 0.33 & 0.25 & 0.17 & 0.75 & 14.89 & 18.87 \\
\bottomrule
\end{tabular}}
\end{table}

\section*{C. MP3DObject Dataset: Construction and Characteristics}

\subsection*{C.1. Construction Pipeline}

MP3DObject is constructed from Matterport3D~\cite{Matterport3D} by extracting object-level point clouds from indoor scenes. For each annotated object instance, we crop the corresponding points, center them, and normalize into a unit sphere, without enforcing any canonical orientation. Extremely incomplete objects whose visible surface area falls below a threshold, as well as degenerate cases with too few points, are discarded. The final dataset contains $4{,}015$ training and $1{,}003$ testing samples over seven shared categories and exhibits substantial variation in layout, occlusion, and pose.

\subsection*{C.2. Qualitative Comparison with Other Datasets}
We qualitatively characterize each existing dataset along four conceptual axes:
\emph{curvature complexity} (how frequently the surface bends or folds),
\emph{extent of missing regions} (size and frequency of holes),
\emph{noise and artifacts} (measurement noise, misalignment, and clutter), and
\emph{pose variability} (degree of non-canonical orientation).

\begin{figure}[htbp]
  \centering
  \includegraphics[width=\linewidth]{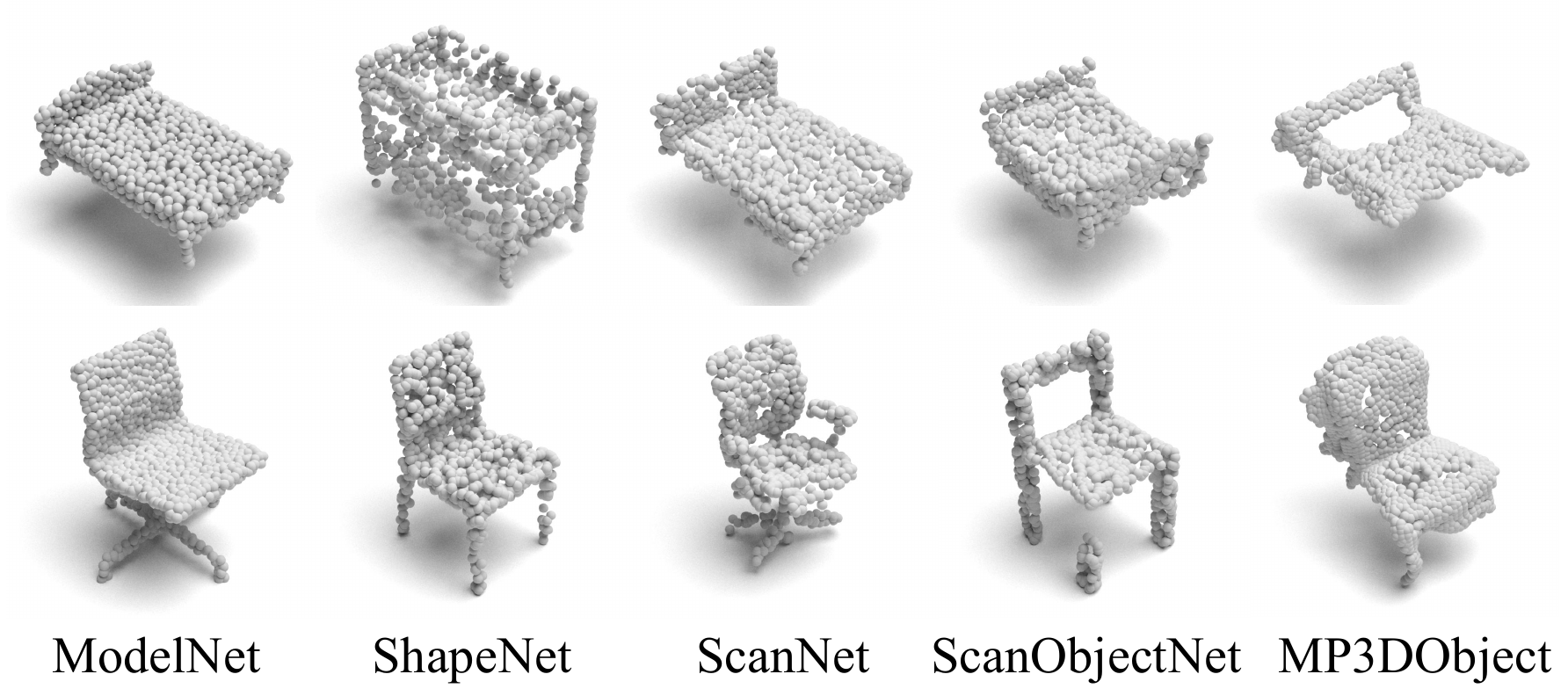}
  \caption{\textbf{Visual comparison of datasets.}
  MP3DObject instances present complex indoor objects in cluttered scenes with substantial occlusions and highly varied poses.
  \emph{For visualization clarity, MP3DObject samples are shown in a manually aligned canonical view; all training and evaluation use the original unaligned scans.}}
  \label{fig:supp_mp3d_vs_others}
\end{figure}

As shown in Table~\ref{tab:supp_dataset_stats} and Figure~\ref{fig:supp_mp3d_vs_others}, MP3DObject tends to have more complex furniture, larger unobserved regions, more cluttered surroundings, and more diverse poses compared to conventional datasets. This makes it a particularly demanding real-scan domain for structure-aware modeling and domain generalization.

\begin{table}[htbp]
\centering
\caption{Qualitative characterization of datasets along four difficulty dimensions. ``Low / mid / high'' indicate relative levels to highlight trends. MP3DObject sits at the most challenging end across all dimensions.}
\label{tab:supp_dataset_stats}
\resizebox{\columnwidth}{!}{
\begin{tabular}{lcccc}
\toprule
Dataset & Curvature & Missing & Noise & Orientation \\
\midrule
ModelNet     & low        & low        & low        & low \\
ShapeNet     & low--mid   & low        & low        & low \\
ScanNet      & mid        & mid--high  & mid        & mid \\
ScanObjectNN & mid--high  & high       & high       & high \\
MP3DObject   & \textbf{high} & \textbf{high} & \textbf{high} & \textbf{high} \\
\bottomrule
\end{tabular}}
\end{table}

\subsection*{C.3. Class-wise Visualization}

\begin{figure*}[htbp]
  \centering
  \includegraphics[width=0.8\textwidth]{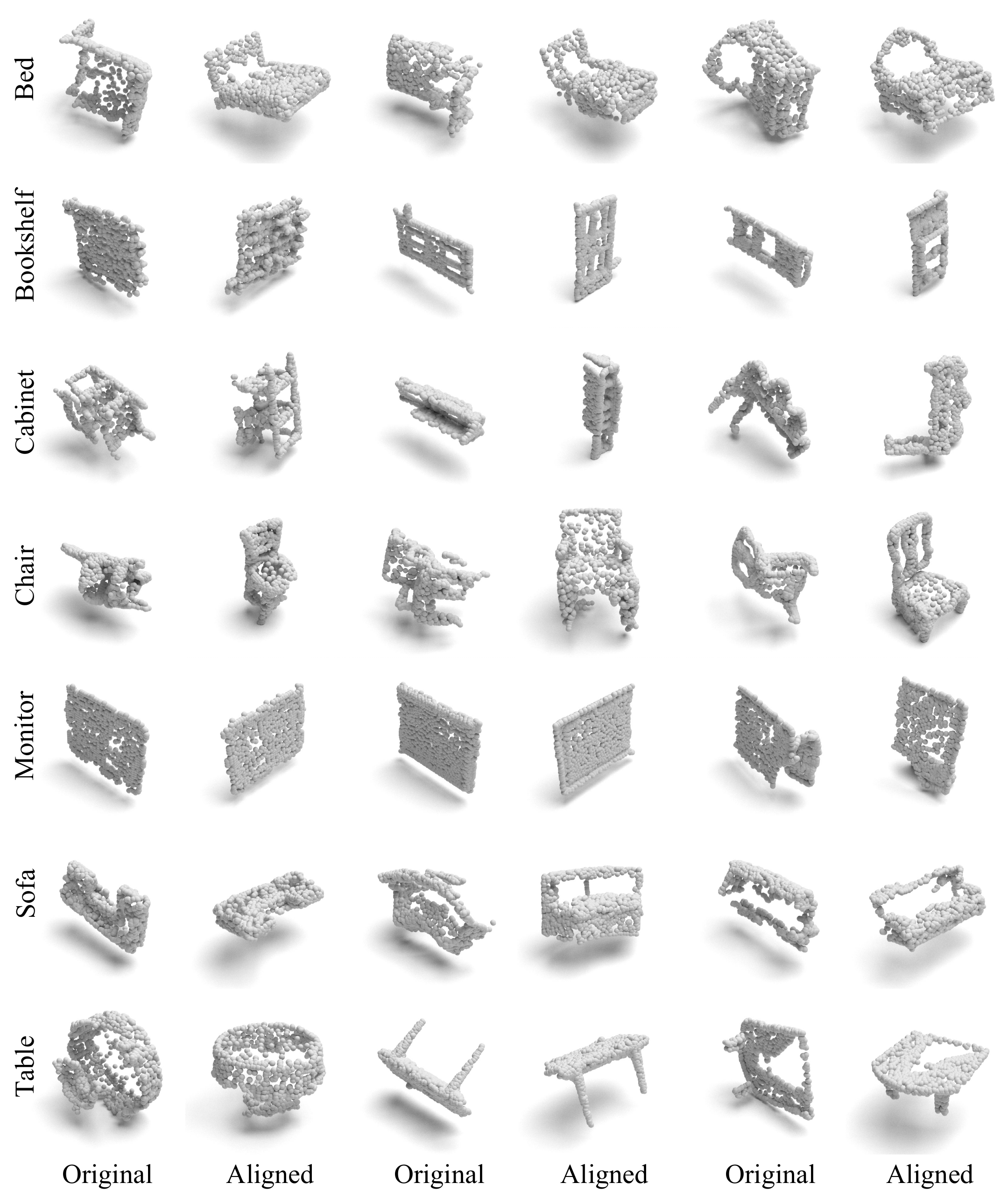}
  \caption{\textbf{MP3DObject per-class visualization in original and aligned poses.}
  For each class, we show several instances in their original unaligned pose (left of each pair) and in a manually aligned pose (right of each pair) to facilitate visual inspection.
  The original pose distribution is highly diverse, reflecting realistic scanning conditions.
  \emph{Alignment is applied \textbf{only} for visualization; all training and evaluation in our experiments use the original unaligned MP3DObject scans.}}
  \label{fig:supp_mp3d_classwise_align}
\end{figure*}

To further illustrate MP3DObject, we provide class-wise visualizations in both the original unaligned pose and a manually aligned pose used only for visualization. For each category (e.g., bed, bookshelf, cabinet, chair, monitor, sofa, table), we randomly select several instances and render them as pairs in Figure~\ref{fig:supp_mp3d_classwise_align}.

\section*{D. Qualitative Evaluation}

\begin{figure*}[htbp]
  \centering
  \includegraphics[width=0.77\textwidth]{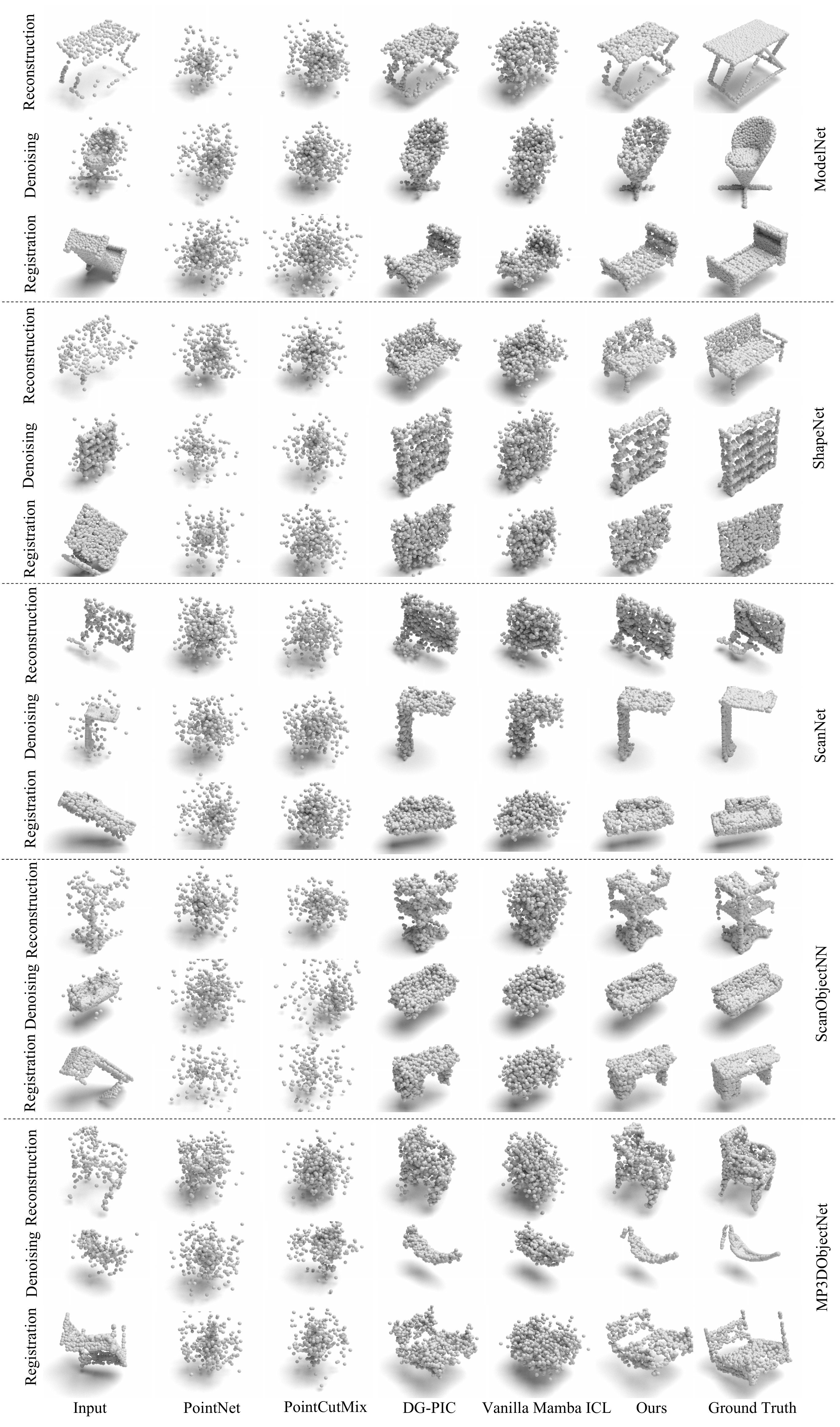}
  \caption{Qualitative comparisons on different target domains.}
  \label{fig:supp_all}
\end{figure*}

We follow the leave-one-domain-out protocol, selecting one dataset as the unseen target and training on the remaining four. Representative qualitative results
for more domains
are shown in Figure~\ref{fig:supp_all}, covering reconstruction, denoising, and registration. The visual patterns follow the quantitative trends in Table~\ref{tab:main_results}. Classical method~\cite{qi2017pointnet} and augmentation-based DG approach methods~\cite{zhang2022pointcutmix} exhibit incomplete recovery of thin structures and occasional topological breaks. DG-PIC~\cite{jiang2024dg} produces more stable outputs but often retains coarse geometry, especially when inputs are severely partial. Coordinate-based Mamba ICL improves long-range consistency but frequently yields fragmented local patches due to its sensitivity to traversal order.

As the target domain becomes more challenging, \textit{i.e.,} 
MP3DObject with complex furniture and large unobserved regions, the qualitative gap widens. Baselines commonly hallucinate missing parts, collapse curved surfaces, or generate disconnected fragments. Across all domains and tasks, our method provides the most coherent reconstructions, smoothest denoising, and most stable registrations, preserving both global structure and fine-grained details. These qualitative observations highlight the benefits of structure-aware serialization, hierarchical domain-aware modeling, and spectral graph alignment under severe domain shift.

\section*{E. Training and Model Hyperparameters}

For reproducibility, we summarize the main training and architectural hyperparameters used in our experiments. These settings correspond to the released codes.

\noindent\textbf{Training configuration.}
We train all models using AdamW with cosine learning rate scheduling:
\begin{itemize}
  \item Optimizer: AdamW, learning rate $1\times10^{-4}$ (for global batch size 96, following linear scaling), weight decay $0.05$, $\beta_1=0.9$, $\beta_2=0.95$, $\epsilon=10^{-8}$.
  \item Scheduler: cosine decay (CosLR) for 300 epochs, with a warmup stage over the first 20 epochs and a minimum learning rate of $10^{-6}$.
  \item Point sampling: each shape is downsampled to 1,024 points for both training and testing.
  \item Batch setting: total batch size 96, gradient accumulation step 1.
  \item Loss: Chamfer Distance L2 (CDL2) for all tasks.
\end{itemize}

\noindent\textbf{Patch and token configuration.}
\begin{itemize}
  \item Number of patches per shape: $G=64$.
  \item Patch size: $S=32$ points per patch.
  \item Patch encoder dimension: 256.
  \item Serialization: default type \texttt{both\_parallel}, applying CDS and GCS in parallel and concatenating their outputs into a sequence of length $4G$.
\end{itemize}

\noindent\textbf{Mamba backbone.}
\begin{itemize}
  \item Token embedding dimension (\texttt{trans\_dim}): 256.
  \item Number of encoder Mamba layers: 4.
  \item Number of decoder Mamba layers: 2.
  \item Bidirectional Mamba: enabled (\texttt{bidir=true}) for all experiments.
  \item Drop path rate: 0.1.
  \item Masking ratio on target segments: 0.7.
  \item Hierarchical Domain-Aware Modeling (HDM): enabled by default.
\end{itemize}

These settings are kept fixed across the experiments reported in the main paper and this supplementary document, except where explicitly varied in ablation studies.

